\documentclass[11pt]{article}

\usepackage[final]{acl}

\usepackage{times}
\usepackage{latexsym}
\usepackage{amssymb}

\usepackage[T1]{fontenc}

\usepackage[utf8]{inputenc}

\usepackage{microtype}

\usepackage{inconsolata}

\usepackage{graphicx}

\usepackage{hyperref}
\usepackage{url}
\usepackage{enumitem}
\usepackage{booktabs}
\usepackage{graphicx}
\usepackage{tabularray}
\usepackage{caption}
\usepackage{graphicx}   %
\usepackage{subcaption} %
\usepackage{xspace}
\usepackage{tcolorbox}
\usepackage{listings}
\usepackage{tcolorbox}      %
\usepackage[table]{xcolor}
\usepackage{amsmath}

\definecolor{mygray}{rgb}{0.94,0.95,0.95}
\lstdefinelanguage{prompt}{
    basicstyle=\normalfont\fontfamily{pcr}\selectfont,
    showstringspaces=false,
    breaklines=True,
    backgroundcolor=\color{mygray},
}

\newcommand{\thinkrow}[1]{\rowcolor{blue!10}  #1 \texttt{</think>} \\}
\newcommand{\searchrow}[1]{\rowcolor{green!10} \texttt{<tool\_call>} #1 \texttt{</tool\_call>} \\}
\newcommand{\resultrow}[1]{\rowcolor{gray!10} \texttt{<tool\_response>} #1 \texttt{</tool\_response>} \\}

\newcommand{\memoryrow}[1]{\rowcolor{yellow!10} \texttt{<memory>} #1 \texttt{</memory>} \\}

\newcommand{\Ours}{MemSearcher\xspace}

\title{MemSearcher: Training LLMs to Reason, Search and Manage Memory via End-to-End Reinforcement Learning}

\author{
  \textbf{Qianhao Yuan}${}^{1,2}$,
  \textbf{Jie Lou}${}^{3}$,
  \textbf{Zichao Li}${}^{1,2}$,
  \textbf{Jiawei Chen}${}^{1,2}$,\\
  \textbf{Yaojie Lu}${}^{1}$\thanks{Corresponding author.},
  \textbf{Hongyu Lin}${}^{1}$,
  \textbf{Le Sun}${}^{1}$,
  \textbf{Debing Zhang}${}^{3}$,
  \textbf{Xianpei Han}${}^{1}$ \\
  ${}^{1}$Chinese Information Processing Laboratory, Institute of Software,\\
  Chinese Academy of Sciences, Beijing, China\\
  ${}^{2}$University of Chinese Academy of Sciences, Beijing, China\\${}^{3}$Xiaohongshu Inc\\
\texttt{\{yuanqianhao2024,lizichao2022,chenjiawei2024\}@iscas.ac.cn} \\
\texttt{\{luyaojie,hongyu,sunle,xianpei\}@iscas.ac.cn  loujie0822@gmail.com} \\
}

\begin{document}
\maketitle
\begin{abstract}

LLM-based search agents often concatenate the full interaction history into the context, producing long and noisy inputs, and increasing compute cost and GPU memory overhead.
To address this issue, we propose MemSearcher, an agent framework that maintains a compact memory during multi-turn interactions, retaining only question-relevant information and thereby keeping the context length stable across turns.
Training MemSearcher is challenging because each trajectory spans multiple turns under different LLM contexts, making each turn an independent optimization target in reinforcement learning. 
We introduce multi-context GRPO,  which propagates trajectory-level advantages to all turns for end-to-end optimization.
Experiments demonstrate that MemSearcher outperforms strong history-concatenation (ReAct-style) baselines on a range of public datasets while maintaining nearly constant token counts across multi-turn interactions.
The code and models will be publicly available at \href{https://github.com/icip-cas/MemSearcher}{https://github.com/icip-cas/MemSearcher}.
\end{abstract}

\section{Introduction}

Recently, Large Language Model (LLM)-based search agents have advanced quickly, delivering substantial improvements on knowledge-acquisition tasks~\citep{jin2025search,chen2025learning,li2025search}.
Unlike traditional Retrieval-Augmented Generation (RAG)~\cite{lewis2020retrieval,yue2024inference,xiong2025rag}, search agents treat search engines as external tools, and autonomously determine when and how to invoke them.
This process typically requires multiple turns of interaction between agents and the environment.

\begin{figure}
    \centering
    \includegraphics[width=\linewidth]{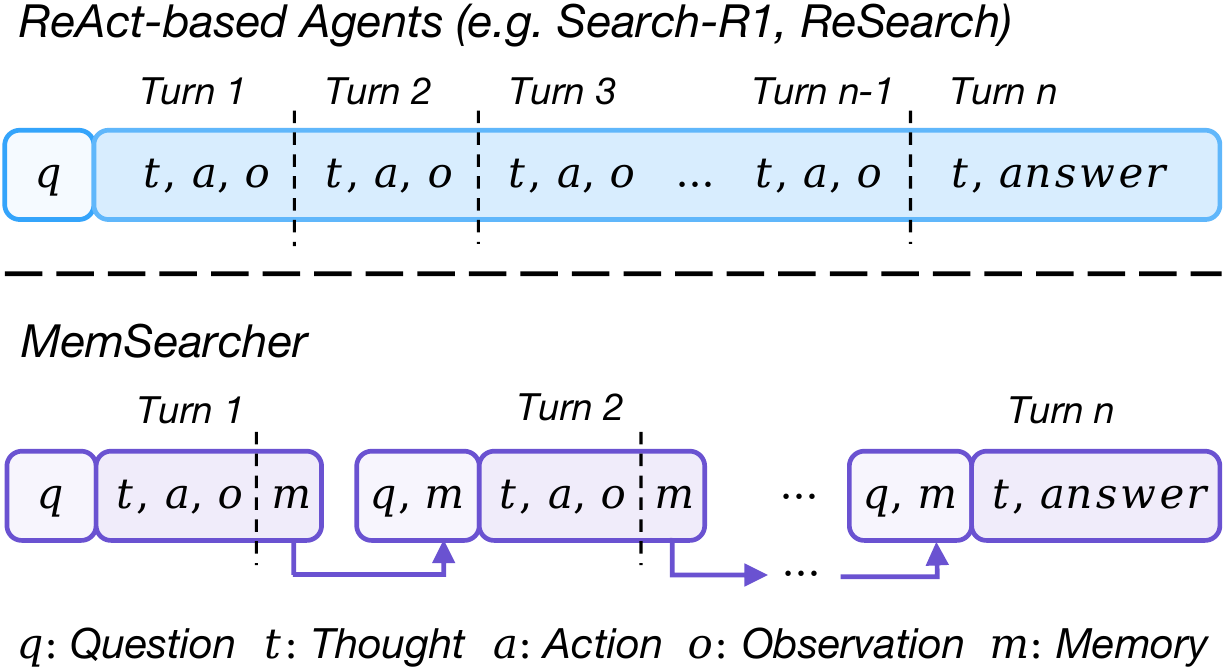}
    \caption{Comparison between ReAct and \Ours.
    ReAct continuously appends all interaction history, including thought $t$, action $a$ and observation $o$ at each turn.
    \Ours iteratively updates a compact memory $m$ that retains only essential information.
    }
    \label{fig:overview}
\end{figure}

A representative paradigm for modeling such multi-turn interactions is ReAct~\citep{yao2023react}.
As shown in Figure~\ref{fig:overview} (Top), a ReAct trajectory is formulated as a multi-turn conversation, where the entire interaction history is incorporated into the context of the backbone LLM.
Although straightforward, this design leads to a fundamental bottleneck: as the number of interaction turns increases, the LLM context grows linearly by appending all thoughts, actions, and observations. 
In the setting of search agents, this results in long and noisy contexts dominated by retrieved passages, many of which are irrelevant to the user’s question. 
Such prolonged contexts not only strain the LLM’s ability to effectively utilize salient information~\cite{liu2023lost,wu2024longmemeval}, but also incur substantial computational and GPU memory overhead. 

In this paper, we introduce \Ours, an agent framework 
that maintains a compact, iteratively updated memory throughout multi-turn interactions, preserving only the information essential for addressing the user's question.
At each turn, \Ours provides the backbone LLM with two succinct inputs, the user's question and the memory, rather than the entire, ever-growing interaction history.
The LLM generates a reasoning trace (thought) and performs an action based on it. 
After the response to the action is returned by the environment, the LLM functions as a memory manager to integrate the memory based on the previous memory and the current interaction.
Since the number of tokens in the memory is restricted by a predefined maximum length, this design keeps LLM contexts stable while preserving salient information throughout multi-turn interactions.

Since current LLMs have not been optimized under the \Ours framework, they are not yet capable of mastering it.
We employ Reinforcement Learning (RL)~\citep{sutton1998reinforcement} to train \Ours agents, which enables the optimization of models without annotated trajectories.
Among RL algorithms, Group Relative Policy Optimization (GRPO)~\citep{shao2024deepseekmath} has emerged as a widely adopted method, as it improves LLM abilities while optimizing the GPU memory usage of Proximal Policy Optimization (PPO)~\citep{schulman2017proximal}.
In \Ours, a trajectory consists of multiple turns under different contexts, making each turn an independent optimization targets.
To enable end-to-end GRPO training of \Ours agents, we introduce multi-context GRPO.
It propagates trajectory-level advantages to each turn, and then optimizes every turn independently, enabling a stable training for \Ours agents.

We conduct extensive experiment across seven public knowledge-acquisition datasets.
\Ours demonstrates superior performance to ReAct-based methods.
Compared to ReAct, which exhibits a linear increase in token numbers with interaction turns, \Ours maintains nearly constant token counts within contexts.
Notably, \Ours supports multi-turn interactions within a compact context window shorter than 4K tokens, enabling deploying search agents in resource-constrained settings.

We summarize our contributions as follows.
\begin{itemize}
    \item We introduce \Ours, an agent framework that leverages the backbone LLM as a memory manager to iteratively retain a compact memory, eliminating the need to append the entire interaction history to LLM contexts.
    \item We propose multi-context GRPO, which provides end-to-end RL training for \Ours trajectories, each of which contains multiple turns under different contexts.
    \item Comprehensive experiments demonstrate that \Ours achieves superior performance to ReAct-based baselines, while maintaining a nearly constant context length across multi-turn interactions.
\end{itemize}

\begin{figure*}
    \centering
    \includegraphics[width=\linewidth]{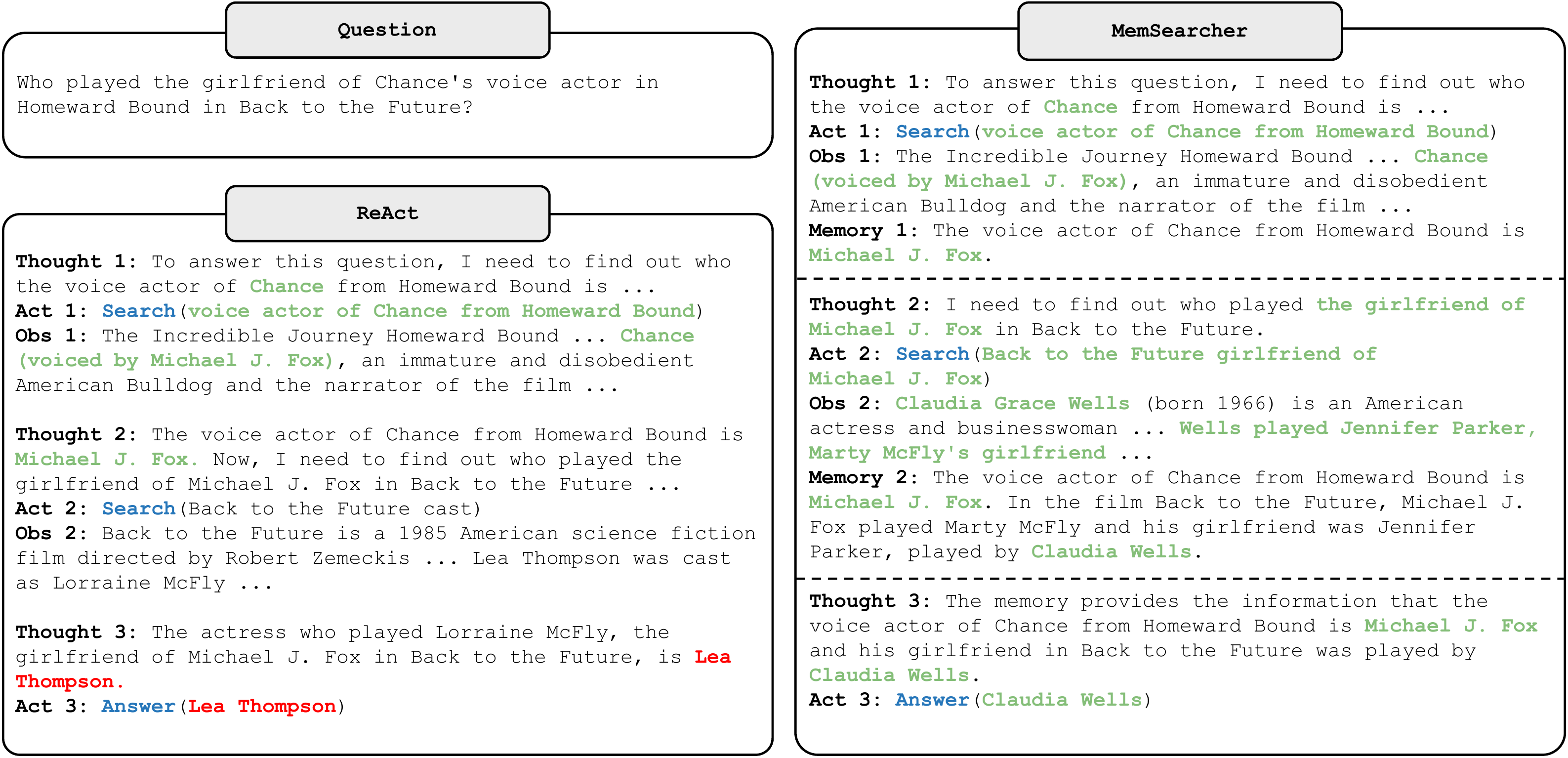} 
    \caption{Comparison between ReAct and \Ours.
    ReAct accumulates all intermediate content in the context, leading to an incorrect answer by confusing the character’s girlfriend with the actor’s on-screen relationship.
    In contrast, MemSearcher maintains an independent context for each turn via a compact memory storing only task-relevant facts, which helps disambiguate entities and relations across turns and yields the correct answer. 
    }
    \label{fig:example}
\end{figure*}

\section{The \Ours Framework}

In this section, we first revisit the ReAct paradigm and its limitation of ever-growing contexts.
Then, we introduce the \Ours framework, which maintains a compact context throughout multi-turn interactions via iterative memory integration.

\subsection{Preliminary: ReAct}
\label{sec:preliminary}

ReAct~\citep{yao2023react} has become the most popular paradigm for building LLM-based agents~\cite{jin2025search, chen2025learning, mo2025livemcpbench}.
The idea of ReAct is simple: a trajectory is a multi-turn conversation, and each turn is an interaction between the agent and the environment.
At each turn, the LLM generates a thought, and performs an action, after which the environment provides an observation in response to the performed action.

Specifically, we assume that at the $i$-th turn, the agent generates thought $t_i$, takes an action $a_i$, and receives an observation $o_i$.
In particular, $o_0=q$ represents the observation prior to the first turn, where $q$ denotes the user's question.
At the $i$-th turn, the input to the LLM is as follows:
\begin{equation}
c_i = (q, t_1, a_1, o_1, \cdots, t_{i-1}, a_{i-1}, o_{i-1}).
\end{equation}
Then, the agent generates $t_i$ and performs the corresponding action $a_i$, following policy $\pi(t_i, a_i|c_i)$.

Although straightforward, ReAct leads to a continuous increase in the number of tokens in the LLM context, due to its design of appending all interaction history.
This increase is linear throughout interactions, placing significant pressure on the inference of LLMs~\cite{hsieh2024ruler,wu2024longmemeval,chen2025consistentchat}.
In addition, in the context of search agents, the observations are passages retrieved by the search engine, which often include substantial noise and information irrelevant to answering the user’s question.
This further constrains the performance of ReAct-based search agents, as illustrated in Figure~\ref{fig:example} (Left).

Moreover, the linear growth in the number of tokens leads to increased computational and GPU memory overhead.
Consequently, more efficient and scalable methods for building search agents need to be explored.

\subsection{Agent with Iterative Memory Integration}
\label{sec:overview}
The \Ours framework is illustrated in Figure~\ref{fig:overview} (Bottom).
At the $i$-th turn, the backbone LLM receives only two succinct inputs: the user's question $q$ between \texttt{<question>} \texttt{</question>}, and a compact memory $m_{i-1}$ expressed in natural language, enclosed within \texttt{<memory>} \texttt{</memory>}.
The memory encapsulates relevant information from previous turns that is considered helpful to answer the question.
In particular, the memory $m_0$ prior to the first turn is empty.
Therefore, the input to the LLM at the $i$-th turn is formulated as:
\begin{equation}
c_i = (q, m_{i-1}).
\end{equation}

\begin{figure*}
    \centering
    \includegraphics[width=\linewidth]{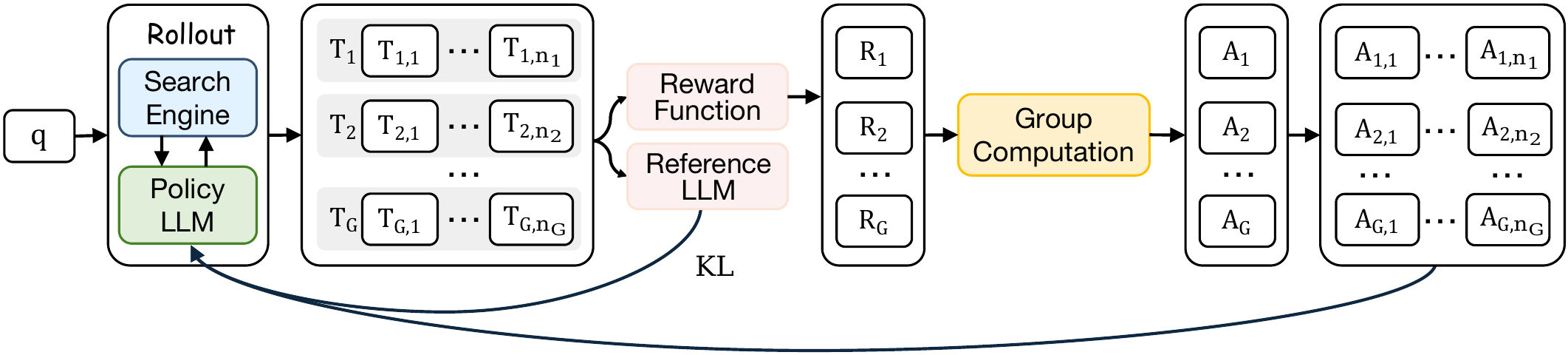}
    \caption{Demonstration of multi-context GRPO.
    In rollout, it first samples a group of trajectories $\{T_i\}_{i=1}^G$ for question $q$.
    The $i$-th trajectory $T_i$ consists of multiple turns $\{T_{i,j}\}_{j=1}^{n_i}$ under different contexts.
    Then, it computes rewards $\{R_i\}_{i=1}^G$, and derives the trajectory-level advantages $\{A_i\}_{i=1}^G$ from these rewards.
    It propagates trajectory-level advantages to each turn within them, \textit{i.e.} $A_{i,j}=A_i$, and treat each turn as an independent optimization target to update the policy LLM.
    }
    \label{fig:grpo}
\end{figure*}

Upon reading the user's query and the previous memory, the LLM generates thought $t_i$ within \texttt{<think>} \texttt{</think>} and performs an action $a_i$ between \texttt{<tool\_call>} \texttt{</tool\_call>}, following policy $\pi(t_i, a_i|c_i)$.
As the action is executed, the environment returns the response $o_i$ within \texttt{<tool\_response>} \texttt{</tool\_response>}.

After receiving $o_i$, \Ours overwrites the previous memory to an updated one for the next turn.
The LLM are asked to carefully read $o_i$ and integrate any new information that helps to answer the question, while preserving all relevant details from the previous memory $m_{i-1}$.
The resulting memory is denoted as $m_i$.
This process continues iteratively, until the maximum number of interactions is reached or sufficient information is gathered and the LLM generates a final answer.

\begin{table}[t]
    \centering
    \resizebox{\linewidth}{!}
    {
    \begin{tabular}{lcccccccc}
        \toprule
        \textbf{Method} & \textbf{Context} & \textbf{FLOPs / Turn} & \textbf{FLOPs} & \textbf{Memory} \\
         \midrule
        ReAct & $O(n)$ & $O(n)$ & $O(n^2)$ & $O(n)$ \\
        \Ours & $O(1)$ & $O(1)$ & $O(n)$ & $O(1)$ \\
        \bottomrule
    \end{tabular}
    }
    \caption{Computational overhead comparison, including tokens in context, FLOPs per turn, total FLOPs and GPU memory, scaling with the number of turns $n$.
    }
    \label{tab:compute}
\end{table}

Unlike ReAct, which continuously  concatenates historical thoughts, actions and observations into the LLM context, \Ours compresses only the essential information into a compact memory, as illustrated in Figure~\ref{fig:example} (Right).

Although the \Ours framework seems simple, it exactly supports the scaling of LLM-based agents.
Since the number of tokens in the memory never exceeds a predefined maximum length, \Ours maintains the context within a few thousands of tokens.
Consequently, the FLOPs per turn keeps $O(1)$, and the total compute scales linearly to the number of turns.
We illustrate the computational overhead comparison between ReAct and \Ours in Table~\ref{tab:compute}.

Specifically, under the setting of search agents with search engines as tools, action $a_i$ takes one of the following two forms: (1) answer to the user's question in $\verb|\boxed{}|$ and terminating the interactions, (2) issuing a search engine call with a query to obtain additional information to answer the query. 
If the latter is chosen, the returned $o_i$ is the relevant passages retrieved from the search engine in response to the search query.

\begin{table*}[t]
    \centering
    \resizebox{\linewidth}{!}
    {
    \begin{tabular}{lcccccccc}
        \toprule
        \textbf{Method} & \textbf{NQ} & \textbf{TriviaQA} & \textbf{PopQA} & \textbf{HotpotQA} & \textbf{2wiki} & \textbf{Musique} & \textbf{Bamboogle} & \textbf{Avg.} \\

         \midrule
        \multicolumn{8}{l}{\textit{Small Size: 3B parameters}} \\
        \midrule
        Search-o1 3B~\cite{li2025search} & 16.6 & 31.0 & 8.2 & 14.8 & 22.4 & 5.2 & 22.4 & 17.2 \\
        Search-R1 3B-base~\cite{jin2025search} & 40.6 & 58.7 & 43.5 & 28.4 & 27.3 & 4.9 & 8.8 & 30.3  \\
        Search-R1 3B-instruct~\cite{jin2025search} & 34.1 & 54.5 & 37.8 & 32.4 & 31.9 & 10.3 & 26.4 & 32.5  \\
        AutoRefine 3B-base~\cite{shi2025search} & \underline{46.7} & \underline{62.0} & \underline{45.0} & \underline{40.5} & \underline{39.3} & 15.7 & 34.4 & \underline{40.5} \\
        AutoRefine 3B-instruct~\cite{shi2025search} & 43.6 & 59.7 & 44.7 & 40.4 & 38.0 & 16.9 & 33.6 & 39.6 \\
        ReSearch 3B~\cite{chen2025learning} & 20.4 & 33.5 & 17.3 & 35.6 & \underline{39.3} & \underline{17.3} & \underline{37.6} & 28.7 \\
        ZeroSearch 3B~\cite{sun2025zerosearch} & 41.4 & 57.4 & 44.8 & 27.4 & 30.0 & 9.8 & 11.1 & 31.7 \\
        O$^2$-Searcher 3B~\cite{mei20252} & 44.4 & 59.7 & 42.9 & 38.8 & 37.4 & 16.0 & 34.4 & 39.1 \\
        \rowcolor{cyan!20} MemSearcher 3B (Ours) & \textbf{47.0} & \textbf{63.8} & \textbf{47.9} & \textbf{43.9} & \textbf{43.5} & \textbf{17.9} & \textbf{42.4} & \textbf{43.8}  \\
        \midrule
        \multicolumn{8}{l}{\textit{Base Size: 7B parameters}} \\
        \midrule
        Search-o1 7B~\cite{li2025search} & 19.4 & 40.6 & 11.4 & 17.0 & 27.0 & 8.6 & 30.4 & 22.1  \\
        Search-R1 7B-base~\cite{jin2025search} & \underline{48.0} & 63.8 & 45.7 & 43.3 & 38.2 & 19.6 & \underline{43.2} & 43.1 \\
        Search-R1 7B-instruct~\cite{jin2025search} & 39.3 & 61.0 & 39.7 & 37.0 & 41.4 & 14.6 & 36.8 & 38.5 \\
        ReSearch 7B~\cite{chen2025learning} & 40.9 & 63.7 & 44.6 & 43.5 & 47.6 & \underline{22.3} & 42.4 & \underline{43.6} \\
        ZeroSearch 7B~\cite{sun2025zerosearch} & 43.6 & \underline{65.2} & \textbf{48.8} & 34.6 & 35.2 & 18.4 & 27.8 & 39.1 \\
        R1-Searcher 7B~\cite{song2025r1} & 40.4 & 52.2 & 41.0 & \underline{44.2} & \textbf{51.3} & 15.8 & 36.8 & 40.2 \\
        \rowcolor{cyan!20} MemSearcher 7B (Ours) & \textbf{52.7} & \textbf{68.1} & \underline{47.8} & \textbf{50.8} & \underline{48.6} & \textbf{25.8} & \textbf{48.8} & \textbf{48.9} \\
        \midrule
        \multicolumn{8}{l}{\textit{Large Size: > 10B parameters}} \\
        \midrule
        Search-o1 14B~\cite{li2025search} & 19.8 & 46.7 & 23.3 & 27.4 & 28.9 & 11.2 & 34.4 & 27.4 \\
        Search-R1 14B-base~\cite{jin2025search} & \underline{48.6} & 67.6 & 48.0 & \underline{46.8} & \underline{47.0} & 24.1 & 52.8 & 47.8 \\
        Search-R1 14B-instruct~\cite{jin2025search} & 42.4 & 66.0 & 44.2 & 43.6 & 37.9 & 21.0 & 48.0 & 43.3 \\
        ReSearch 32B~\cite{chen2025learning} & 45.5 & \underline{69.4} & \underline{48.2} & 46.7 & 44.9 & \underline{26.4} & \underline{56.8} & \underline{48.3} \\
        \rowcolor{cyan!20} MemSearcher 14B (Ours) & \textbf{53.7} & \textbf{71.1} & \textbf{48.8} & \textbf{51.8} & \textbf{51.5} & \textbf{27.2} & \textbf{57.6} & \textbf{51.7}  \\
        \bottomrule
    \end{tabular}
    }
    \caption{Performance comparison. 
    The best performance is highlighted in \textbf{bold}, while the second-best performance is indicated with an \underline{underline}. 
    }
    \label{tab:main}
\end{table*}

\section{\Ours Training}

In this section, we introduce the end-to-end RL training method of \Ours, including multi-context GRPO and the reward modeling.

\subsection{Multi-Context GRPO}

We use reinforcement learning (RL) to train \Ours agents, since it allows models to evolve without annotated trajectories.
Among RL algorithms, we utilize Group Relative Policy Optimization (GRPO)~\citep{shao2024deepseekmath}, as it optimizes the GPU memory usage of Proximal Policy Optimization (PPO)~\citep{schulman2017proximal}.

Vanilla GRPO samples a group of trajectories $\{T_1, T_2, \cdots, T_G\}$ for each query $q$, and optimizes policy $\pi_{\theta}$ by maximizing the following objective:
\begin{equation}
\scriptsize
\begin{split}
\mathcal{J}(&\theta)
= \mathbb{E}_{q\sim D,\{T_{i}\}_{i=1}^G \sim \pi_{\theta_{\text{old}}}(\cdot|q)}
\frac{1}{G}\sum_{i=1}^G
\Big[\min\bigl(\frac{\pi_\theta(T_i |q)}{\pi_{\theta_{old}}(T_i |q)} A_i,\\
&\qquad\mathrm{clip}\bigl(\frac{\pi_\theta(T_i |q)}{\pi_{\theta_{old}}(T_i |q)},1-\epsilon,1+\epsilon\bigr) A_i\bigr)
- \beta\,\mathbb{D}_{KL}\!\bigl(\pi_{\theta}\,\|\,\pi_{ref}\bigr)\Big],
\end{split}
\label{eq:GRPO-obj}
\end{equation}
where $A_i$ represents the advantage, calculated with the rewards $\{R_1, R_2, \cdots, R_G\}$ within each group:
\begin{equation}
    A_i = \frac{R_i - {\mathrm mean(\{R_1, R_2, \cdots, R_G\})}}{{\mathrm std(\{R_1, R_2, \cdots, R_G\})}}.
\label{eq:advantage}
\end{equation}

For \Ours agents, a trajectory consists of multiple turns under different LLM contexts, each of which is an independent optimization target for LLM RL training.
Therefore, we introduce multi-context GRPO to enable end-to-end GRPO training for \Ours, as illustrated in Figure~\ref{fig:grpo}.

Specifically, we assume that trajectory $T_i$ contains $n_i$ turns, represented as $\{T_{i,1}, T_{i,2}, \cdots, T_{i,n_i}\}$.
According to Section~\ref{sec:overview}, the $j$-th turn can be formulated as:
\begin{equation}
  T_{i,j} = 
  \begin{cases}
    (q, m_{i,j-1}, t_{i,j}, a_{i,j}, o_{i,j}, m_{i,j}), & \text{if $j$$<$$n_i$} \\
    (q, m_{i,j-1}, t_{i,j}, a_{i,j}), & \text{if $j$$=$$n_i$} \\
  \end{cases}
\end{equation}
where $m_{i,j-1}$ is the previous memory, $m_{i,j}$ is the new generated memory, $t_{i,j}$, $a_{i,j}$ and $o_{i,j}$ are the reasoning trace, performed action and tool response, respectively.

We compute reward $R_i$ for each trajectory, and calculate its advantage $A_i$ within the group by Equation~\ref{eq:advantage}.
Then, we uniformly propagate this advantage to all turns within the trajectory, and use each turn as an independent target to optimize the policy model.
The training objective is:
\begin{equation}
\small
\begin{split}
\mathcal{J}(\theta)
= &\mathbb{E}_{q\sim D,{\color{red}\{T_{i,j}\}_{i=1}^G \sim \pi_{\theta_{\text{old}}}(\cdot|c_{i,j})}}
\frac{1}{\color{red}\sum_{i=1}^G n_i}\sum_{i=1}^G{\color{red}\sum_{j=1}^{n_i}} \Big[\\
&\min\bigl({\color{red}r_{i,j}(\theta)} A_{i,j}, \mathrm{clip}\bigl({\color{red}r_{i,j}(\theta)},\,1-\epsilon,\,1+\epsilon\bigr)\,A_{i,j}\bigr)\\
&- \beta\,\mathrm{KL}\!\bigl(\pi_{\theta}\,\|\,\pi_{\text{ref}}\bigr)\Big],
\end{split}
\end{equation}
where $c_{i,j}=(q,m_{i,j-1})$ is the input to the model,
\begin{equation}
  r_{i,j}(\theta)=\frac{\pi_\theta(T_{i,j} |c_{i,j})}{\pi_{\theta_{old}}(T_{i,j} |c_{i,j})} \text{ and } A_{i,j}=A_i.
\end{equation}

Notably, $t_{i,j}$ consists of tokens from both the policy model and the environment.
Following previous RL-based search agents~\citep{jin2025search,chen2025learning}, we use loss mask for tokens from the search engine, ensuring the policy gradient objective is computed only over model-generated tokens, thereby stabilizing RL training.

\subsection{Reward Modeling}
Similar to DeepSeek-R1~\citep{guo2025deepseek}, our reward function considers two parts: format reward and answer reward.
\begin{itemize}[leftmargin=*]
  \item \textbf{Format Reward}: It checks whether the rollout correctly follows the predefined format, including the correctness of usage of tags and the existence of $\verb|\boxed{}|$ in the answer.
  \item \textbf{Answer Reward}: A rule-based reward assesses the correctness of the model's response. It is calculated by using the F1 score between the final answer inside $\verb|\boxed{}|$ and the ground truth.
\end{itemize}

The reward function is formulated as:
\begin{equation}
  R = 
  \begin{cases}
  0, & \text{incorrect format}, \\
  0.1, & \text{correct format \& F1 score$=$0}, \\
    \text{F1 score}, & \text{correct format \& F1 score$>$0}. \\
  \end{cases}
\end{equation}

\section{Experiments}
\label{experiments}

To verify the effectiveness of \Ours, we conduct experiments on a range of datasets and different scales of LLMs.

\subsection{Experiment Setups}
\label{sec:setups}

\paragraph{Baselines.} 
We compare \Ours against multiple baseline methods, including: 
(1) prompt-based multi-step inference with search, such as Search-o1~\citep{li2025search}; (2) LLMs with search trained via Reinforcement Learning, such as Search-R1~\citep{jin2025search}, ReSearch~\citep{chen2025learning}, AutoRefine~\citep{shi2025search}, ZeroSearch~\citep{sun2025zerosearch}, O$^2$-Searcher~\cite{mei20252}, and R1-Searcher~\citep{song2025r1}.

\begin{table*}[t]
    \centering
    \resizebox{0.95\linewidth}{!}
    {
    \begin{tabular}{llcccccccc}
        \toprule
       \textbf{Model} & \textbf{Method} & \textbf{NQ} & \textbf{TriviaQA} & \textbf{PopQA} & \textbf{HotpotQA} & \textbf{2wiki} & \textbf{Musique} & \textbf{Bamboogle} & \textbf{Average} \\
         \midrule
       Qwen2.5-3B & \textit{w/o} training & 16.4 & 23.8 & 22.5 & 11.9 & 11.0 & 3.7 & 11.2 & 14.4  \\
       -Instruct & \textit{w/} training & \textbf{47.0} & \textbf{63.8} & \textbf{47.9} & \textbf{43.9} & \textbf{43.5} & \textbf{17.9} & \textbf{42.4} & \textbf{43.8}  \\
        \midrule
       Qwen2.5-7B & \textit{w/o} training & 22.1 & 41.2 & 23.5 & 27.4 & 27.8 & 11.6 & 27.2 & 25.8  \\
       -Instruct & \textit{w/} training & \textbf{52.7} & \textbf{68.1} & \textbf{47.8} & \textbf{50.8} & \textbf{48.6} & \textbf{25.8} & \textbf{48.8} & \textbf{48.9} \\
        \midrule
       Qwen2.5-14B & \textit{w/o} training & 26.2 & 53.7 & 33.2 & 28.6 & 24.2 & 10.1 & 17.6 & 27.7  \\
       -Instruct & \textit{w/} training & \textbf{53.7} & \textbf{71.1} & \textbf{48.8} & \textbf{51.8} & \textbf{51.5} & \textbf{27.2} & \textbf{57.6} & \textbf{51.7}  \\
        \bottomrule
    \end{tabular}
    }
    \caption{Comparison between models with and without RL training.
    }
    \label{tab:training}
\end{table*}

Among these baselines, ZeroSearch and R1-Searcher interact with the realistic web environment via Google Web Search in their evaluation.

\begin{figure}[t]
    \centering
    \includegraphics[width=\linewidth]{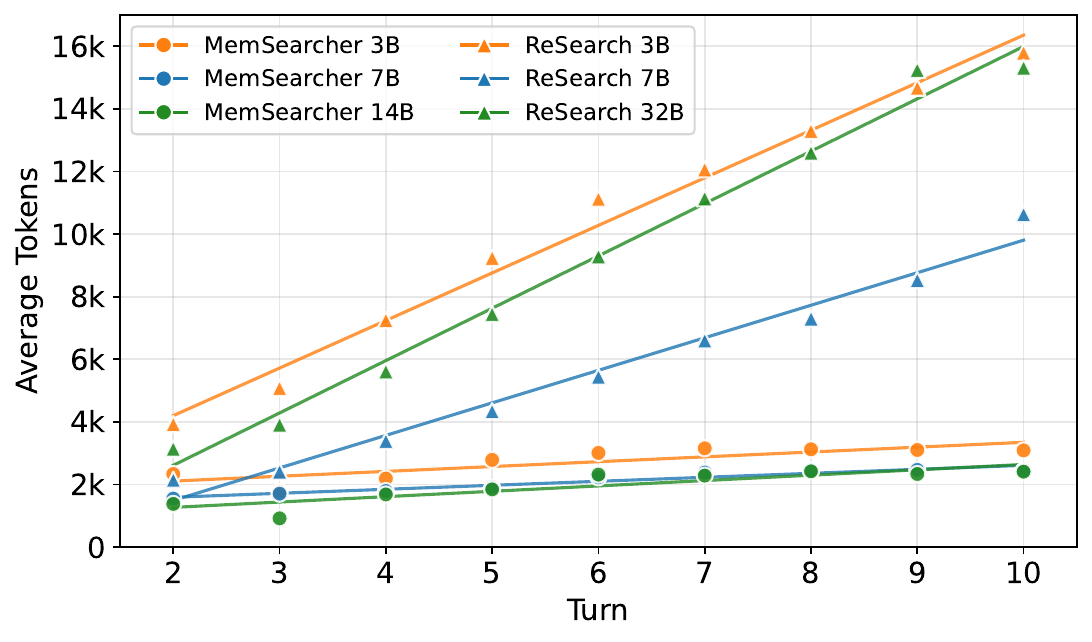}
    \caption{Comparison of the average token numbers in the LLM context between \Ours and ReAct-based ReSearch~\cite{chen2025learning}. 
    }
    \label{fig:tokens}
\end{figure}

\paragraph{Benchmarks and Evaluation Metrics.} We compare \Ours with the baselines across a range of public benchmarks that encompass search with reasoning challenges, such as
Natural Questions (NQ)~\citep{kwiatkowski2019natural}, TriviaQA~\citep{joshi2017triviaqa}, PopQA~\citep{mallen2022not},
Bamboogle~\citep{press2022measuring}, Musique~\citep{trivedi2022musique}, HotpotQA~\citep{yang2018hotpotqa}, and 2WikiMultiHopQA~\citep{ho2020constructing}.
Following Search-R1~\cite{jin2025search}, we use Exact Match (EM) as the evaluation metric.

\paragraph{Implementation Details.} We conduct our experiments on Qwen2.5-3B/7B/14B-Instruct models~\cite{qwen2025qwen25technicalreport}.
We use the 2018 Wikipedia dump~\citep{karpukhin2020dense} as the knowledge source and E5~\citep{wang2022text} as the retriever.
The maximum number of tokens in the memory is set to 1,024.
For training, we follow Search-R1, using its fully open training datasets, including the training splits of NQ~\citep{kwiatkowski2019natural} and HotpotQA~\citep{yang2018hotpotqa}, and the rollout group size of each prompt is set to 5.
We conduct the training based on the verl library~\citep{sheng2025hybridflow}, and the rollout temperature is set to 1.0.
For evaluation, we test \Ours on seven public datasets, covering both in-domain and out-of-domain scenarios.
Such a comprehensive evaluation provides deeper insights into the effectiveness of \Ours under varied conditions.

\begin{table*}[t]
    \centering
    \resizebox{0.85\linewidth}{!}
    {
    \begin{tabular}{lcccccccc}
        \toprule
        \textbf{Method} & \textbf{NQ} & \textbf{TriviaQA} & \textbf{PopQA} & \textbf{HotpotQA} & \textbf{2wiki} & \textbf{Musique} & \textbf{Bamboogle} & \textbf{Average} \\
         \midrule
        SFT & 26.0 & 34.1 & 29.3 & 23.2 & 26.6 & 13.9 & \textbf{46.4} & 28.5  \\
        RL & \textbf{47.0} & \textbf{63.8} & \textbf{47.9} & \textbf{43.9} & \textbf{43.5} & \textbf{17.9} & 42.4 & \textbf{43.8}  \\
        \bottomrule
    \end{tabular}
    }
    \caption{Comparison between models trained via SFT and RL, based on Qwen2.5-3B-Instruct.
    }
    \label{tab:sft}
\end{table*}

\begin{figure*}[t]
  \centering
  \begin{subfigure}{0.4\linewidth}
    \includegraphics[width=\linewidth]{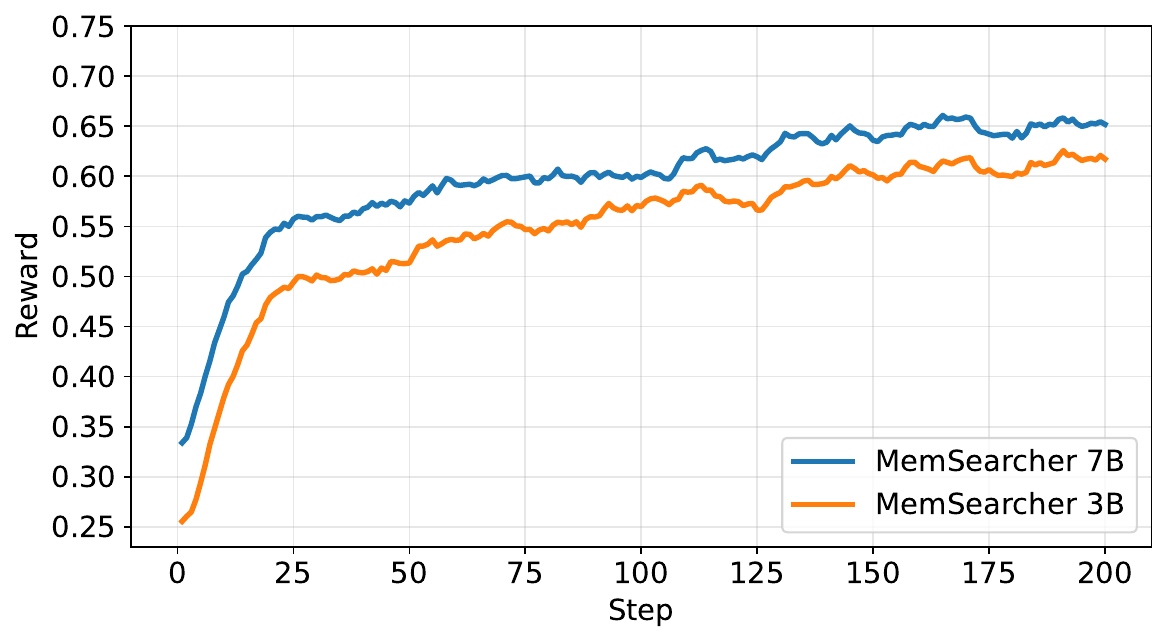}
    \caption{Training Reward}
  \end{subfigure}
  \hspace{0.04\linewidth}
  \begin{subfigure}{0.4\linewidth}
    \includegraphics[width=\linewidth]{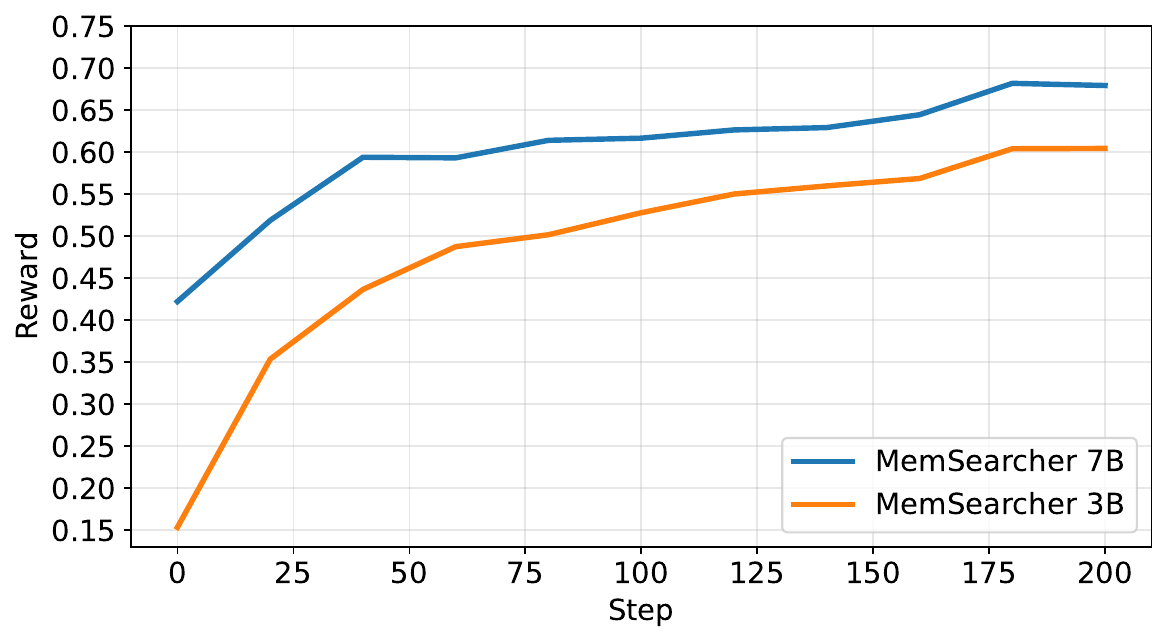}
    \caption{Validation Reward}
  \end{subfigure}
  \caption{Training and validation reward. The validation is conducted on a part of development set of HotpotQA with 100 randomly selected samples, and conducted every 20 steps. The curves are smoothed for clarity.
  }
  \label{fig:reward}
\end{figure*}

\subsection{Main Results}
\subsubsection{Performance Analysis}
In Table~\ref{tab:main}, we provide a comprehensive performance comparison between \Ours and the baselines across the evaluated benchmarks.
Several key observations can be drawn from these results: 

(1) \textbf{\Ours consistently outperforms baselines}, 
demonstrating the effectiveness of our method. 
These performance improvements are consistently observed across both in-distribution datasets such as NQ and HotpotQA, and out-of-distribution datasets, such as TriviaQA, PopQA, 2WikiMultiHopQA, Musique and Bamboogle.

(2) \textbf{Remarkably, even with a smaller LLM size, \Ours outperforms larger baseline models.}
For example, \Ours 3B attains an average score of 43.8 accross the seven benchmarks, surpassing the baselines with 7B parameters.
Moreover, \Ours 7B achieves an average score of 48.9, even higher than ReSearch 32B.
These results suggest that \Ours makes more effective use of model capacity.

(3) \textbf{\Ours surpasses the baselines that rely on the realistic web search engine.} 
Specifically, \Ours achieves superior performance compared to ZeroSearch and R1-Searcher, both of which depend on Google Web Search to retrieve external information during their evaluation.

\subsubsection{Efficiency Analysis}
In addition to the performance gains, \Ours achieves superior efficiency compared to ReAct-based agents, as it eliminates the need to include all interaction history within the LLM context.

To validate this, we record the number of tokens in the LLM contexts of \Ours and ReAct-based ReSearch at each turn and calculate their average across the evaluated benchmarks.
The results are illustrated in Figure~\ref{fig:tokens}.

\textbf{Compared to ReSearch, which exhibits an linear increase in token consumption during the interaction process, \Ours maintains substantially lower and more stable token counts. }
This efficiency gain is primarily attributed to the design of \Ours, which iteratively updates a memory as context, preserving only the essential information for answering the given  question throughout interactions.
Notably, \Ours enables multi-turn interactions within a compact context window shorter than 4K tokens.
Together with the observation that smaller \Ours models even outperform larger baseline models, these results demonstrate the feasibility of deploying search agents in resource-constrained settings.

\subsection{Further Analysis}
\subsubsection{Do we need RL Training?}
To investigate the impact of RL training on the performance of \Ours, we perform a comparative analysis.
The baselines are Qwen2.5-3B/7B/14B-Instruct models, both of which are integrated with the \Ours framework but do not undergo the RL training.

As shown in Table~\ref{tab:training}, the models without RL training demonstrate a pronounced performance degradation across all evaluated benchmarks.
This observation highlights the necessity of RL training in equipping models with the ability to effectively interact with both the search engine and memory, thereby enhancing their overall functionality and task-solving ability.

\paragraph{Why use RL for \Ours instead of SFT?}
Because SFT relies on token-level supervision and require explicit annotations of intermediate memory states, which are costly. Moreover, many larger LLMs also have not been optimized under the \Ours framework, making them suboptimal teachers in distillation-based SFT. 
In contrast, RL directly rewards behaviors that lead to correct final answers, enabling models to learn what to retain and discard during interactions.

As shown in Table~\ref{tab:sft}, we compare the models trained via SFT and RL.
For SFT, we adopt the distillation mechanism. 
Specifically, questions from the RL training data are used to prompt Qwen2.5-72B-Instruct, and trajectories with correct final answers are collected to fine-tune Qwen2.5-3B-Instruct.
We can observe that the model trained via RL demonstrates superior performance compared to the model with SFT training.

\subsubsection{Training and Validation Reward}
We present the curves of training and validation reward in Figure~\ref{fig:reward}, which offer an intuitive view of the learning dynamics during training.
For validation, we construct a validation dataset by randomly sampling 100 examples from the development set of HotpotQA.
We conduct validation every 20 training steps.
The observed reward patterns reveals the following two phases of learning: 

(1) Early stage (first 25 steps). The reward increases sharply, indicating that the models rapidly acquire the fundamental ability to interact effectively with the search engine and memory.

(2) Later stage (after 25 steps). The reward grows at a more gradual pace.
This improvement suggests that the models are refining their strategy, progressively enhancing its capacity to exploit the search engine and manage memory.

The difference between the two stages underscores the transition from basic skill acquisition to the optimization of more advanced behaviors.

\begin{figure}[t]
  \centering

  \begin{subfigure}[t]{0.48\linewidth}
    \centering
    \includegraphics[width=0.75\linewidth]{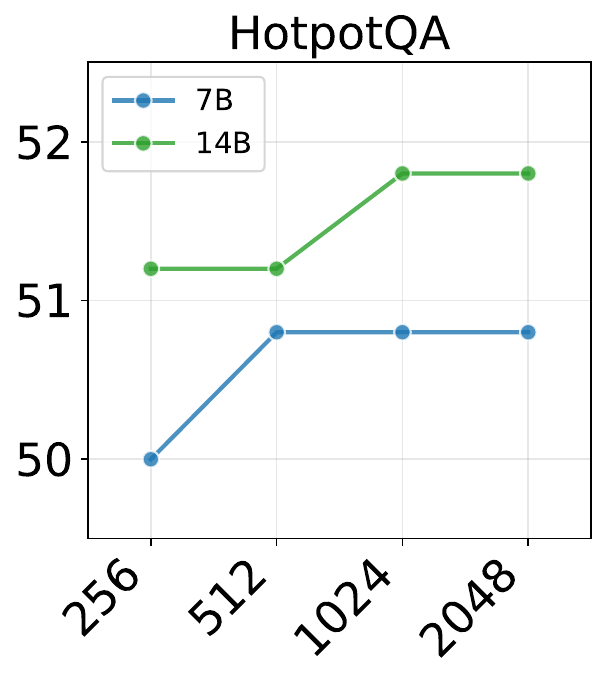}
  \end{subfigure}
  \begin{subfigure}[t]{0.48\linewidth}
    \centering
    \includegraphics[width=0.75\linewidth]{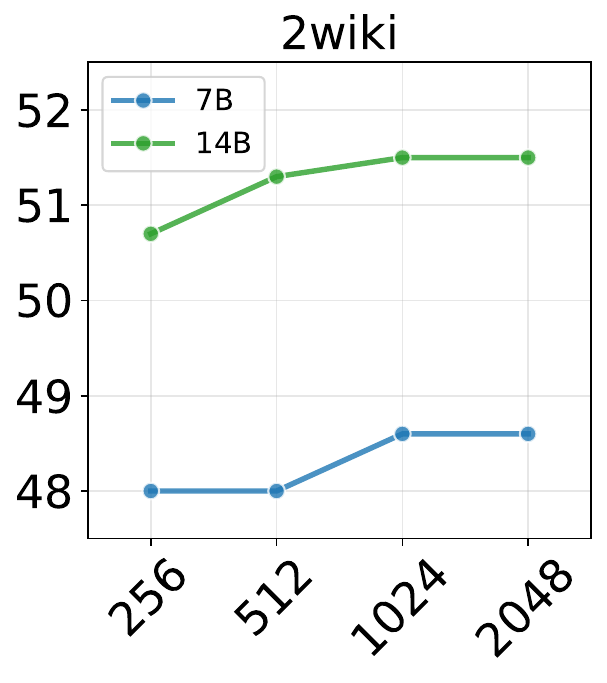}
  \end{subfigure}

  \begin{subfigure}[t]{0.48\linewidth}
    \centering
    \includegraphics[width=0.75\linewidth]{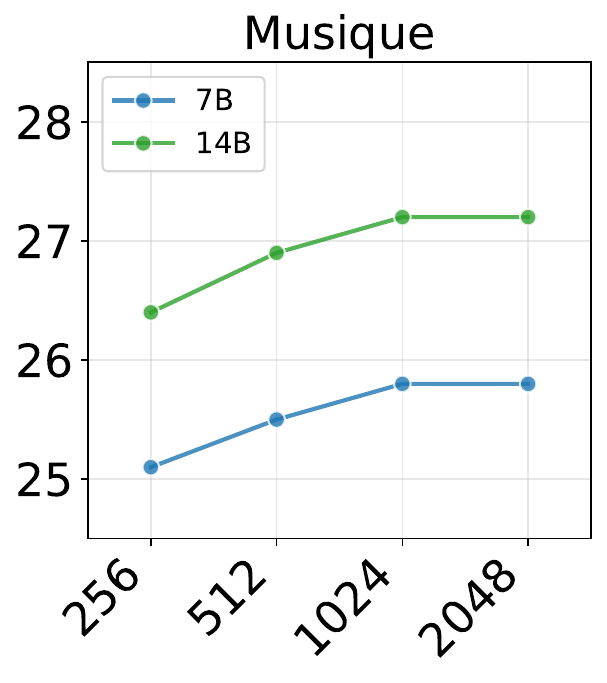}
  \end{subfigure}
  \begin{subfigure}[t]{0.48\linewidth}
    \centering
    \includegraphics[width=0.75\linewidth]{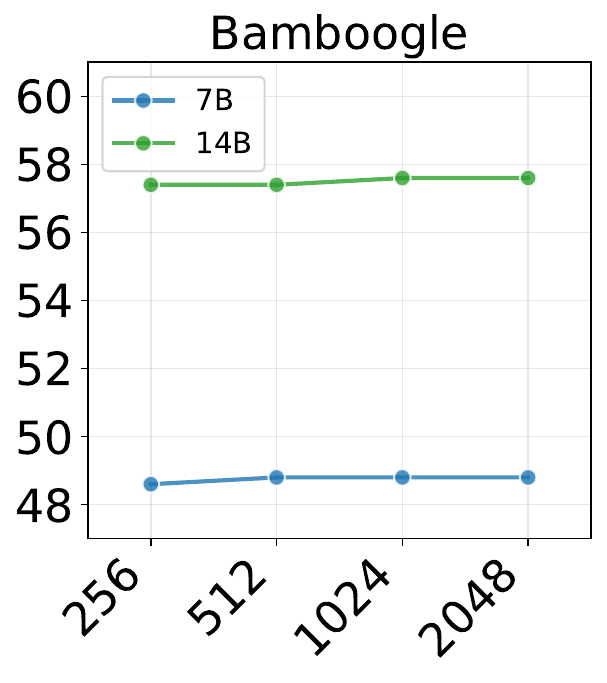}
  \end{subfigure}

  \caption{Ablation on maximum memory lengths.}
  \label{fig:length}
\end{figure}

\subsubsection{Impact of Maximum Memory Length}
Choosing an appropriate maximum memory length for \Ours agents entails a fundamental trade-off between information capacity and memory efficiency. 
A larger maximum length enables agents to preserve more information throughout interactions, but it also exacerbates redundancy. 
In contrast, an overly restrictive memory limit forces the agents to discard essential information, undermining multi-turn search and reasoning performance.
In our default configuration, we set the maximum memory length to 1,024 tokens.

To systematically examine the impact of this hyperparameter, we conduct an ablation study with maximum memory lengths ranging from 256 to 2,048 tokens. 
As shown in Figures~\ref{fig:length}, the results suggest that intermediate memory sizes provide the most favorable trade-off: they retain sufficient task-relevant information while limiting redundancy in memory integration. 
On relatively simple datasets such as Bamboogle, \Ours already achieves superior and saturated performance with a maximum memory length of 256 tokens, suggesting that \Ours is able to compress essential information into a small number of tokens.
In contrast, on more complex datasets such as Musique, model performance improves steadily as the maximum memory length increases from 256 to 1,024 tokens.

\section{Related Work}

\subsection{LLMs with Search Engines}
Large Language Models (LLMs) have demonstrated strong and robust capabilities in solving complex tasks~\citep{openai2024gpt4technicalreport,comanici2025gemini, men2025shortgpt, yuan2025saisa, yuan2025shortv, zeng2025glm}, yet often lack knowledge in specific domains~\citep{peng2023study, li2023large,zhang2025siren, zheng2026deeppresenter}.
Conventional solutions like Retrieval-Augmented Generation (RAG)~\citep{lewis2020retrieval, yue2024inference,yu2022generate} address this by augmenting prompts with retrieved documents.
Alternatively, search agents formulate the problem as an interactive process, treating search engines as external tools.
ReAct~\citep{yao2023react} integrates search into the reasoning process by interleaving it with Chains-of-Thought (CoT)~\citep{wei2022chain} steps.
Recent studies~\citep{jin2025search, chen2025learning} develop agentic reinforcement learning (RL) to enhance search agents.
However, current RL-based approaches~\citep{wu2025webdancer, tao2025webshaper} predominantly adhere to the ReAct paradigm, lacking the exploration of more efficient paradigms.

\subsection{Context Management}

Although context management is pivotal for long-horizon tasks, most LLM agents utilize ReAct~\citep{yao2023react}, which leads to prolonged token sequences. To address this, various memory mechanisms have been proposed: (1) RAG-style memory~\citep{jimenez2024hipporag, zhong2024memorybank} treats memory as an external knowledge source to be retrieved when needed~\citep{zhu2023ghost}. (2) Token-level memory~\citep{jin2024disentangling, zhou2025mem1, orlicki2025beyond} uses SFT or RL algorithms~\citep{schulman2017proximal} to regulate contexts. Strategies range from maintaining latent tokens~\citep{wang2025m+}, forget-resistant buffers~\citep{yang2024memory3}, or to compressing long-context inputs into summaries~\citep{yu2025memagent}. (3) Structured memory~\citep{zeng2024structural,zhang2025g,chhikara2025mem0} organizes information into structured representations, such as the knowledge graphs utilized in Zep~\citep{rasmussen2025zep} and the atomic memory units in A-MEM~\cite{xu2025mem}.
In this paper, we utilize the backbone LLM of the search agent as an intrinsic memory manager, optimizing it via end-to-end multi-context GRPO.

\section{Conclusion}
In this paper, we propose \Ours, an agent framework that integrates a memory with the backbone LLM throughout interactions between the agent and the environment.
It eliminates the need to append all historical thoughts, actions and observations within the LLM context, as in ReAct.
We further introduce multi-context GRPO to optimize \Ours-based agents in an end-to-end fashion.
\Ours demonstrate superior performance across a range of public benchmarks compared to ReAct-based baselines.
Moreover, \Ours maintains nearly constant context length shorter than 4K tokens throughout multi-turn interactions with the environment.

\section*{Limitations}
Although \Ours demonstrates superior performance and efficiency compared to ReAct-based methods, there are several limitations.
While \Ours adopts a simple and effective memory design, more sophisticated memory mechanisms, such as RAG-like memory or structured memory, need further investigation for RL-based search agents.
In addition, Multi-context GRPO serves as an effective extension of the vanilla GRPO algorithm, yet more techniques need to be explored to mitigate potential length bias and to exploit the potential of RL training.

\section*{Acknowledgment}
We sincerely thank the reviewers for their insightful comments and valuable suggestions. This work was supported by the National Key R\&D Program of China (2024YFC3308000), the Natural Science Foundation of China (No. 62536008, 62306303, 62476265).

\bibliography{acl2026_conference}

\appendix
\section{Appendix}
\subsection{LLM Usage}
We used OpenAI’s ChatGPT to help polish the language and improve the readability of the manuscript. Specifically, ChatGPT was used for grammar checking and sentence rephrasing.
We list our prompt for using OpenAI’s ChatGPT to help polish writing as follows.

\begin{tcolorbox}[title=Prompt for Using OpenAI’s ChatGPT to Help Polish Writing]
\begin{lstlisting}[language=prompt, escapeinside={(*@}{@*)}]
Below is a paragraph from an academic paper. Polish the writing to meet the academic style,improve the spelling, grammar, clarity, concision and overall readability. Furthermore, list all modification and explain the reasons to do so in markdown table. \\
Paragraph: {(*@\textcolor{red}{paragraph}@*)}
\end{lstlisting}
\end{tcolorbox}

\subsection{Training Hyperparameters}
\label{appendix:training}
We train \Ours agents with full parameter optimization and gradient checkpointing.
We show training hyperparameters in Table~\ref{tab:implementation-details}.
The 3B/7B experiments are run on 8 H100 GPUs and 14B on 2×8 H100 GPUs.

\begin{table}[htbp]
\begin{center}
\begin{tabular}{l|l}
\toprule
\textbf{Parameter} & \textbf{Value} \\
\midrule
Learning Rate & 1e-6 \\
Train Batch Size & 256 \\
Number of Training Epochs & 1 \\
Number of Rollout & 5 \\
Rollout Temperature & 1.0 \\
KL Loss Coefficient & 0.001 \\
Clip Ratio & 0.2 \\
\bottomrule
\end{tabular}
\end{center}
\caption{Training details of \Ours.}
\label{tab:implementation-details}
\end{table}

\subsection{Details of Evaluated Datasets}
We evaluate \Ours agents on the following public question answering datasets:
\begin{itemize}[leftmargin=0.4cm]
    \item \textbf{Natural Questions (NQ)}~\citep{kwiatkowski2019natural}, a QA dataset with questions consisting of real anonymized, aggregated queries issued to the Google search engine. 
    \item \textbf{TriviaQA}~\citep{joshi2017triviaqa}, a large scale challenging dataset with relatively complex, compositional questions, requireing more reasoning to find answers.
    \item \textbf{PopQA}~\citep{mallen2022not}, a dataset consisting of 14k questions covering factual information in the long tail.
    \item \textbf{Bamboogle}~\citep{press2022measuring}, a manually constructed dataset with multi-hop questions, where all questions are sufficiently difficult to be unanswerable by a single search engine call.
    \item \textbf{Musique}~\citep{trivedi2022musique}, a QA dataset with 25K questions that by require proper multi-hop reasoning.
    \item \textbf{HotpotQA}~\citep{yang2018hotpotqa}, a dataset with 113k Wikipedia-based question-answer pairs, where the questions requires finding and reasoning over multiple supporting documents to answer, and and not constrained to any existing knowledge bases or knowledge schemas.
    \item \textbf{2WikiMultiHopQA}~\citep{ho2020constructing}, a multi-hop QA dataset that uses both structured and unstructured data.
    It is challenging for multi-hop models and explicitly ensures that multi-hop reasoning is required.
\end{itemize}

\begin{table*}[htbp]
  \begin{center}
  \begin{tabular}{p{\linewidth}}
  \toprule
  \textbf{Question:} What was the form of the language that the last name Sylvester comes from, used in the era of Rotrude's father, later known as? \\
  \textbf{Answer:} Medieval Latin \\
  \midrule 
   \thinkrow{\textbf{Turn 1:} \texttt{<think>} To answer this question, I need to find out which language the last name Sylvester comes from. Let's start by searching Wikipedia for this information.}
  \searchrow{\{"name": "wikipedia\_search", "arguments": \{"query": "Last name Sylvester comes from what language"\}\}}
  \resultrow{Sylvester James Jr. (September 6, 1947 - December 16, 1988), who used the stage name of Sylvester, was an American singer-songwriter. Primarily active in ... ...
  }
  \memoryrow{The last name Sylvester comes from the Latin language.}
  \midrule 
  \thinkrow{\textbf{Turn 2:} \texttt{<think>} The memory mentions that the last name Sylvester comes from the Latin language. However, it does not provide any information about the form of the Latin language used in the era of Rotrude's father, later known as. I need to find out which era Rotrude's father lived in.}
  \searchrow{\{"name": "wikipedia\_search", "arguments": \{"query": "Who was Rotrude's father, and what was he later known as?"\}\}}
  \resultrow{Rotrude Rotrude (or sometimes referred to as Hruodrud/Hruodhaid) (775/778 – 6 June 810) was a Frankish princess, the second daughter of Charlemagne from his marriage to Hildegard. Few clear records ... ...
  }
  \memoryrow{The last name Sylvester comes from the Latin language. Rotrude was a Frankish princess, the second daughter of Charlemagne from his marriage to Hildegard.}
  \midrule 
  \thinkrow{\textbf{Turn 3:} \texttt{<think>} The memory mentions that the last name Sylvester comes from the Latin language. However, it does not provide any information about the form of the Latin language used in the era of Rotrude's father, later known as. I need to find out who Charlemagne was.}
  \searchrow{\{"name": "wikipedia\_search", "arguments": \{"query": "Who was Charlemagne"\}\}}
  \resultrow{Charlemagne or Charles the Great (2 April 742 – 28 January 814), numbered Charles I, was King of the Franks from 768, King of the Lombards from 774, and Holy Roman Emperor from ... ...
  }
  \memoryrow{The last name Sylvester comes from the Latin language. Rotrude was a Frankish princess, the second daughter of Charlemagne from his marriage to Hildegard. Charlemagne was later known as Holy Roman Emperor.}
  \midrule 
  \thinkrow{\textbf{Turn 4:} \texttt{<think>} The last name Sylvester comes from the Latin language. Charlemagne was later known as Holy Roman Emperor. 
The question asks what was the form of the Latin language used in the era of Rotrude's father, later known as. 
Since Rotrude's father was Charlemagne, and Charlemagne was later known as Holy Roman Emperor, we need to find out what was the form of the Latin language used in his era. 
I will use the wikipedia\_search function to find this information. }
\searchrow{\{"name": "wikipedia\_search", "arguments": \{"query": "Form of Latin language used in the era of Charlemagne"\}\}}
\resultrow{Medieval Latin was the form of Latin used in Roman Catholic Western Europe during the Middle Ages ... ... 
}
\memoryrow{The last name Sylvester comes from the Latin language. Rotrude was a Frankish princess, the second daughter of Charlemagne from his marriage to Hildegard. Charlemagne was later known as Holy Roman Emperor. 
The form of Latin language used in the era of Charlemagne was Medieval Latin.}
\midrule
\thinkrow{\textbf{Turn 5:} \texttt{<think>} The memory mentions that the form of Latin language used in the era of Charlemagne was Medieval Latin.}
The answer to the question is $\verb|\boxed{Medieval Latin}|$.
  \\
  \bottomrule
  \end{tabular}
  \end{center}
  \caption{A case study of \Ours.}
  \label{tab:case-study}
  \end{table*}

\subsection{Case Study}
To provide a more intuitive understanding of the utility of \Ours, we present a case study in Table~\ref{tab:case-study}.
The text enclosed by \texttt{<think>} and \texttt{</think>}, \texttt{<tool\_call>} and \texttt{</tool\_call>}, as well as \texttt{<memory>} and \texttt{</memory>} is generated by the model.
The text enclosed by \texttt{<tool\_response>} and \texttt{</tool\_response>} is retrieved from the search engine.
This case demonstrates that the model can effectively maintain a compact memory, retaining only the essential information necessary to solve the question.

\end{document}